\PassOptionsToPackage{unicode}{hyperref}
\PassOptionsToPackage{hyphens}{url}
\PassOptionsToPackage{dvipsnames,svgnames,x11names}{xcolor}
\documentclass[
]{article}

\usepackage{amsmath,amssymb}
\usepackage{iftex}
\ifPDFTeX
  \usepackage[T1]{fontenc}
  \usepackage[utf8]{inputenc}
  \usepackage{textcomp} % provide euro and other symbols
\else % if luatex or xetex
  \usepackage{unicode-math}
  \defaultfontfeatures{Scale=MatchLowercase}
  \defaultfontfeatures[\rmfamily]{Ligatures=TeX,Scale=1}
\fi
\usepackage{lmodern}
\ifPDFTeX\else  
\fi
\IfFileExists{upquote.sty}{\usepackage{upquote}}{}
\IfFileExists{microtype.sty}{% use microtype if available
  \usepackage[]{microtype}
  \UseMicrotypeSet[protrusion]{basicmath} % disable protrusion for tt fonts
}{}
\makeatletter
\@ifundefined{KOMAClassName}{% if non-KOMA class
  \IfFileExists{parskip.sty}{%
    \usepackage{parskip}
  }{% else
    \setlength{\parindent}{0pt}
    \setlength{\parskip}{6pt plus 2pt minus 1pt}}
}{% if KOMA class
  \KOMAoptions{parskip=half}}
\makeatother
\usepackage{xcolor}
\ifx\paragraph\undefined\else
  \let\oldparagraph\paragraph
  \renewcommand{\paragraph}[1]{\oldparagraph{#1}\mbox{}}
\fi
\ifx\subparagraph\undefined\else
  \let\oldsubparagraph\subparagraph
  \renewcommand{\subparagraph}[1]{\oldsubparagraph{#1}\mbox{}}
\fi

\providecommand{\tightlist}{%
  \setlength{\itemsep}{0pt}\setlength{\parskip}{0pt}}\usepackage{longtable,booktabs,array}
\usepackage{calc} % for calculating minipage widths
\usepackage{etoolbox}
\makeatletter
\patchcmd\longtable{\par}{\if@noskipsec\mbox{}\fi\par}{}{}
\makeatother
\IfFileExists{footnotehyper.sty}{\usepackage{footnotehyper}}{\usepackage{footnote}}
\makesavenoteenv{longtable}
\usepackage{graphicx}
\makeatletter
\def\maxwidth{\ifdim\Gin@nat@width>\linewidth\linewidth\else\Gin@nat@width\fi}
\def\maxheight{\ifdim\Gin@nat@height>\textheight\textheight\else\Gin@nat@height\fi}
\makeatother
\setkeys{Gin}{width=\maxwidth,height=\maxheight,keepaspectratio}
\makeatletter
\def\fps@figure{htbp}
\makeatother
\newlength{\cslhangindent}
\newlength{\csllabelwidth}
\newlength{\cslentryspacingunit} % times entry-spacing
\newenvironment{CSLReferences}[2] % #1 hanging-ident, #2 entry spacing
 {% don't indent paragraphs
  \setlength{\parindent}{0pt}
  \ifodd #1
  \let\oldpar\par
  \def\par{\hangindent=\cslhangindent\oldpar}
  \fi
  \setlength{\parskip}{#2\cslentryspacingunit}
 }%
 {}
\usepackage{calc}

\usepackage{booktabs}
\usepackage{longtable}
\usepackage{array}
\usepackage{multirow}
\usepackage{wrapfig}
\usepackage{float}
\usepackage{colortbl}
\usepackage{pdflscape}
\usepackage{tabu}
\usepackage{threeparttable}
\usepackage{threeparttablex}
\usepackage[normalem]{ulem}
\usepackage{makecell}
\usepackage{xcolor}
\usepackage{multicol}
\usepackage{hhline}
\newlength\Oldarrayrulewidth
\newlength\Oldtabcolsep
\usepackage{hyperref}
\usepackage{arxiv}
\usepackage{orcidlink}
\usepackage{amsmath}
\usepackage[T1]{fontenc}
\makeatletter
\@ifpackageloaded{caption}{}{\usepackage{caption}}
\AtBeginDocument{%
\ifdefined\contentsname
  \renewcommand*\contentsname{Table of contents}
\else
  \newcommand\contentsname{Table of contents}
\fi
\ifdefined\listfigurename
  \renewcommand*\listfigurename{List of Figures}
\else
  \newcommand\listfigurename{List of Figures}
\fi
\ifdefined\listtablename
  \renewcommand*\listtablename{List of Tables}
\else
  \newcommand\listtablename{List of Tables}
\fi
\ifdefined\figurename
  \renewcommand*\figurename{Figure}
\else
  \newcommand\figurename{Figure}
\fi
\ifdefined\tablename
  \renewcommand*\tablename{Table}
\else
  \newcommand\tablename{Table}
\fi
}
\@ifpackageloaded{float}{}{\usepackage{float}}
\floatstyle{ruled}
\@ifundefined{c@chapter}{\newfloat{codelisting}{h}{lop}}{\newfloat{codelisting}{h}{lop}[chapter]}
\floatname{codelisting}{Listing}

\makeatother
\makeatletter
\@ifpackageloaded{caption}{}{\usepackage{caption}}
\@ifpackageloaded{subcaption}{}{\usepackage{subcaption}}
\makeatother
\makeatletter
\@ifpackageloaded{tcolorbox}{}{\usepackage[skins,breakable]{tcolorbox}}
\makeatother
\makeatletter
\@ifundefined{shadecolor}{\definecolor{shadecolor}{rgb}{.97, .97, .97}}
\makeatother
\ifLuaTeX
  \usepackage{selnolig}  % disable illegal ligatures
\fi
\IfFileExists{bookmark.sty}{\usepackage{bookmark}}{\usepackage{hyperref}}
\IfFileExists{xurl.sty}{\usepackage{xurl}}{} % add URL line breaks if available
\hypersetup{
  pdftitle={Efficient Multi-domain Text Recognition Deep Neural Network Parameterization with Residual Adapters},
  pdfauthor={Jiayou Chao; Wei Zhu},
  pdfkeywords={Deep Neural Network, Optical Character
Recognition, Multi-domain Adapter, Multi-task Learning, Continual
Learning},
  colorlinks=true,
  linkcolor={blue},
  filecolor={Maroon},
  citecolor={Blue},
  urlcolor={Blue},
  pdfcreator={LaTeX via pandoc}}

\newcommand{\runninghead}{A Preprint }
\title{Efficient Multi-domain Text Recognition Deep Neural Network
Parameterization with Residual Adapters}
\author{
\textbf{Jiayou Chao}~\orcidlink{0000-0003-4410-5309}\\\\Stony Brook
University\\\\\href{mailto:jiayou.chao@stonybrook.edu}{jiayou.chao@stonybrook.edu}\\\\\\
\textbf{Wei Zhu}\\\\Stony Brook
University\\\\\href{mailto:wei.zhu@stonybrook.edu}{wei.zhu@stonybrook.edu}}
\date{}
\begin{document}
\maketitle
\begin{abstract}
Recent advancements in deep neural networks have markedly enhanced the
performance of computer vision tasks, yet the specialized nature of
these networks often necessitates extensive data and high computational
power. Addressing these requirements, this study presents a novel neural
network model adept at optical character recognition (OCR) across
diverse domains, leveraging the strengths of multi-task learning to
improve efficiency and generalization. The model is designed to achieve
rapid adaptation to new domains, maintain a compact size conducive to
reduced computational resource demand, ensure high accuracy, retain
knowledge from previous learning experiences, and allow for
domain-specific performance improvements without the need to retrain
entirely. Rigorous evaluation on open datasets has validated the model's
ability to significantly lower the number of trainable parameters
without sacrificing performance, indicating its potential as a scalable
and adaptable solution in the field of computer vision, particularly for
applications in optical text recognition.
\end{abstract}
{\bfseries \emph Keywords}
\def\sep{\textbullet\ }
Deep Neural Network \sep Optical Character Recognition \sep Multi-domain
Adapter \sep Multi-task Learning \sep 
Continual Learning

\ifdefined\Shaded\renewenvironment{Shaded}{\begin{tcolorbox}[breakable, borderline west={3pt}{0pt}{shadecolor}, boxrule=0pt, interior hidden, enhanced, sharp corners, frame hidden]}{\end{tcolorbox}}\fi

\hypertarget{introduction}{%
\section{Introduction}\label{introduction}}

As deep neural networks continue to dramatically improve results for
nearly all traditional computer vision problems, the community has begun
to shift its focus to more ambitious objectives (LeCun, Bengio, and
Hinton 2015). One prevalent pragmatic constraint associated with deep
neural networks is related to their notable propensity for
specialization towards a singular task, as well as their substantial
requirements in terms of data size and computational resources (Ruder
2017; Reeve, Cannings, and Samworth 2021). This holds particularly true
for the most efficacious deep neural networks, which are commonly
trained on extensive datasets comprising millions of images. This issue
poses a challenge in numerous applications where the available data is
constrained and the computational resources are limited (Li et al. 2018;
Cannings, Fan, and Samworth 2019). An additional limitation of this
method is its lack of scalability, particularly when confronted with an
increasing number of problems to be resolved. Furthermore, this approach
lacks efficiency due to the repetitive acquisition of the same
information and the inability of models to transfer knowledge across
different tasks (Misra et al. 2016; S. Liu et al. 2022; S. Liu, Johns,
and Davison 2019; Sinha et al. 2020; Rothenhäusler and Bühlmann 2023).
The concept of employing a singular model to address multiple tasks is
highly attractive due to its capacity to facilitate the transfer of
acquired knowledge from one task to another. The significance of this
matter is particularly pronounced when considering tasks that exhibit
interrelatedness, such as object detection and segmentation, or object
detection and classification. In this particular scenario, the acquired
knowledge from one task can be effectively utilized to enhance the
performance of the other task.

The increasing focus on developing data representations that exhibit
robust performance across diverse problem domains and datasets is indeed
noteworthy. The realization that such adaptable representations are
essential for developing machine learning systems that can effectively
generalize beyond the constrictions of particular tasks and datasets is
what is driving this burgeoning interest. As the field of artificial
intelligence continues to evolve, the ability to create models that can
seamlessly adapt and maintain high levels of accuracy across a variety
of challenges has become a pinnacle pursuit for researchers and
practitioners alike (Parisi et al. 2019; L. Wang et al. 2023). Most of
the works in this area focus on image classification (Bhattacharjee,
Süsstrunk, and Salzmann 2023; S. Liu et al. 2022; Han Zhao et al. 2018;
Peng et al. 2019; Ganin et al. 2016; Misra et al. 2016; Rebuffi,
Vedaldi, and Bilen 2018) or text classification (P. Liu, Qiu, and Huang
2017; Ganin et al. 2016; Guo, Pasunuru, and Bansal 2020; Z. Wang et al.
2021; Houlsby et al. 2019), yet their application in optical character
recognition (OCR) remains somewhat unexplored, to the best of our
knowledge. The incorporation of multi-task learning in OCR offers
substantial benefits, particularly in practical applications. A key
aspect of OCR is the unique and valuable information each entry
provides, which can significantly enhance recognition accuracy and speed
(Veeramachaneni and Nagy 2003; Mathis and Breuel 2002; Ho and Nagy
2000). For example, when digitalizing a business form, recognizing an
entry as a phone number immediately implies that it contains only
numerical digits. This context-specific insight is crucial for improving
recognition accuracy and efficiency. Similarly, when processing entries
in foreign languages, incorporating language-specific information can
substantially reduce recognition errors. Leveraging domain-specific
knowledge further refines the accuracy of OCR systems. Therefore, a
versatile OCR model adept at utilizing domain-specific information is
immensely beneficial for a wide array of real-world OCR scenarios,
underscoring the value of such an approach.

The innovation introduced in this research paper is a multi-domain
neural network architecture designed specifically for enhancing OCR
across diverse applications. This architecture capitalizes on the
concept of dynamic adaptability, employing adapter modules that function
as interstitial components within the established neural network
framework. These modules serve as vectors for domain-specific
parameters, strategically integrated within a preexisting, pre-trained
model to fine-tune its feature extraction capabilities to suit new
tasks. The introduction of adapter modules into the neural network is a
strategic response to the issue of catastrophic forgetting, a problem
where sequential learning of new tasks can lead to a degradation of
performance on previously learned tasks (French 1999; Goodfellow et al.
2015; Kirkpatrick et al. 2017). By preserving the adapters corresponding
to previous domains intact, the network maintains its proficiency across
all learned tasks. The architecture thus proposes a scalable solution
that promotes efficient adaptation without compromising historical
knowledge. The intricacies of the proposed methodology necessitate
precise domain specification for optimal feature extraction during data
input. In instances where the domain remains ambiguous, an ancillary
neural network is suggested as a viable mechanism for domain prediction,
before processing by the primary OCR-focused architecture (Bengio,
Courville, and Vincent 2014).

This innovative approach is in line with the wider discussion on
transfer learning and domain adaptation in neural networks. It builds
upon the fundamental research that suggests the usefulness of
fine-tuning for specific tasks in deep learning models (Rebuffi,
Vedaldi, and Bilen 2018; Houlsby et al. 2019). In addition, the ongoing
inquiry enhances the discourse by presenting a pragmatic and adaptable
framework that can be easily adjusted for a wide range of OCR
applications, thus marking a significant contribution to the field of
machine learning and text recognition. To summarize, the paper presents
a robust, multi-domain neural network architecture enhanced through the
integration of domain-adaptive modules. This system is aimed at
addressing the challenges of catastrophic forgetting and domain
specificity in OCR, and it offers a versatile framework adaptable to an
evolving array of tasks.

The efficacy of the proposed multi-domain neural network architecture
was rigorously assessed using publicly available datasets, offering a
transparent and replicable benchmark for the evaluation process. The
experiments underscored the model's proficiency in striking a balance
between model complexity and performance. Notably, the architecture
demonstrated a marked reduction in the number of trainable
parameters---indicative of an efficient parameterization---without
compromising the integrity of its OCR capabilities. The results affirm
the model's potential as a scalable and adaptable solution for OCR
challenges across a multitude of domains.

\hypertarget{related-work}{%
\section{Related Work}\label{related-work}}

Training a deep learning model for multi-domains or general purposes has
long been the focus of academic research. The major topics usually
include multi-task learning, adapting new domains and avoiding
forgetting.

\textbf{Multi-task learning} (MTL) aims at learning multiple related
tasks simultaneously by sharing information and computation among them.
Early work (Caruana 1997) in this area focuses on deep neural network
(DNN) models which share weights in the earlier layers and use
specialized ones in the later layers. This line of research focuses on
learning a diverse set of tasks in the same visual domain. In this case,
the knowledge learned for one task can be used to improve the
performance of the other one. However, this approach usually requires
the different tasks to be related to each other, and to share the same
input data. If the problem setting is an input distribution \(p(X)\),
then the goal of MTL is to learn a single model \(f(X)\) that can be
used to address multiple different tasks \(T_1, T_2, \dots, T_n\), where
the label distribution \(p(Y_i|X)\) is different for each task. In this
case, the model \(f(X)\) is a function of the input \(X\) and the task
\(T_i\), and it outputs the label \(Y\).

\textbf{Sequential Learning} (Incremental Learning or Life-long
Learning), is a theoretical framework that aims to acquire a model for a
substantial number of tasks sequentially, while retaining the
information acquired from previous tasks. Notably, this approach assumes
that the data from previous tasks are no longer accessible during the
training of subsequent tasks. Catastrophic forgetting is possible in
sequential learning, albeit it does not usually occur (Parisi et al.
2019; Hanbin Zhao et al. 2021). There is no guarantee that it can retain
prior knowledge. If sometimes one is interested in maximizing the
performance on a specific new task, sequential learning can be used as a
form of initialization for the new task. In this case, the model is
trained on the old tasks, and then it is fine-tuned on the new task.
This approach is called Transfer Learning (TL).

\textbf{Progressive Learning} is yet another concept to solve complex
sequences of tasks. This approach excels in leveraging knowledge
transfer while avoiding catastrophic forgetting, distinguishing it from
traditional methods. As elucidated by Rusu et al. (2022), Progressive
Learning models are uniquely designed to be immune to forgetting and
efficiently utilize prior knowledge through lateral connections to
previously learned features. The general procedure of a Progressive
Learning model begins with training a deep learning model on an initial
task. Upon completion, the model's weights are frozen to preserve the
learned knowledge. Subsequently, a new model is trained on a second
task, with its weights also being frozen post-training. Critically, the
weights of this second model are interconnected with the first model's
weights via lateral connections, facilitating knowledge transfer and
feature integration. This process is iteratively applied to each
subsequent task, culminating in a final model that amalgamates the
knowledge acquired from all tasks. Fayek, Cavedon, and Wu (2020)
highlights the effectiveness of this method in maintaining a robust
knowledge base across multiple tasks.

Unlike finetuning, which primarily utilizes prior knowledge at the
initialization phase, Progressive Networks retain a reservoir of
pretrained models. These models are then employed to learn lateral
connections, thereby extracting and integrating useful features for new
tasks. This strategy fosters a rich compositionality where previously
acquired knowledge is consistently integrated at every layer of the
feature hierarchy. Additionally, the integration of new capacities
alongside pre-trained networks endows these models with the flexibility
to both repurpose old computations and assimilate new ones. This
contrasts with finetuning, where the model is constrained to either
completely reuse or disregard old computations. However, the scalability
of Progressive Networks presents a significant challenge. As the number
of tasks increases, the model's parameters grow exponentially, which
poses limitations for practical applications. This issue, highlighted in
recent studies, suggests a need for innovative approaches to manage
parameter growth efficiently. Moreover, the implementation of
Progressive Learning models, particularly in learning from a sequence of
tasks, demands intricate design and execution strategies. The efficacy
of these models is highly contingent on the optimal sequencing of task
execution, a challenge that remains an active area of research.

\textbf{Adapters} serve as a lightweight alternative to complete model
fine-tuning, as they involve the introduction of a small collection of
parameters specifically at each backbone layer. Adapters address many
constraints commonly encountered in the process of complete model
fine-tuning. They exhibit advantages such as parameter efficiency,
accelerated training iterations, and the ability to be shared and
combined owing to their modular and compact nature. In addition, it is
worth noting that adapters typically provide comparable performance to
the current leading approach of full fine-tuning (Hu et al. 2023;
Rohanian et al. 2023; Hanbin Zhao et al. 2021; Pfeiffer et al. 2020;
Rücklé et al. 2021).

The utilization of adapters offers various benefits, particularly in
terms of parameter efficiency. The number of adapter parameters that
have been updated constitutes only a small proportion of the overall
parameters in the fully fine-tuned model. This feature facilitates the
easy sharing, storage, and dynamic loading of adapters, hence enhancing
adaptability and versatility across various jobs. Moreover, the compact
nature of adapters, along with the fact that most of the weights are
frozen, makes them a computationally efficient option for fine-tuning.
This holds especially true for tasks that necessitate continuous
retraining.

\hypertarget{sec-method}{%
\section{Method}\label{sec-method}}

The proposed framework delineates an innovative Convolutional Recurrent
Neural Network (CRNN) architecture that is holistically trainable and
comprises a sophisticated feature extraction network augmented with
adapter modules, in addition to a sequential network. The fundamental
component of the feature extraction network is a convolutional neural
network that draws inspiration from the ResNet architecture introduced
by He et al. (2016), specifically engineered to distill features from
the input imagery. This network diverges from the quintessential ResNet
by the incorporation of residual adapters after each stacked residual
block within the feature extraction network. These residual adapters,
inspired by the work of Rebuffi, Bilen, and Vedaldi (2017), are
constituted by a vector of \(1 \times 1\) convolutional filter banks
functioning in concert with an identity skip connection, and are
tailored to fine-tune the extracted features to various tasks.

The sequential aspect of the network employs a transformer model, a
construct that excels in encoding sequential information (Vaswani et al.
2023). This section of the network is further enhanced by bottleneck
adapters, an innovation introduced by Houlsby et al. (2019), which are
situated subsequent to the multi-head attention and feed-forward layers
in the transformer. These adapters are notable for their limited number
of parameters relative to the attention and feedforward layers prevalent
in traditional models. They also feature a skip-connection, enhancing
the efficiency of training.

\begin{figure}

{\centering \includegraphics{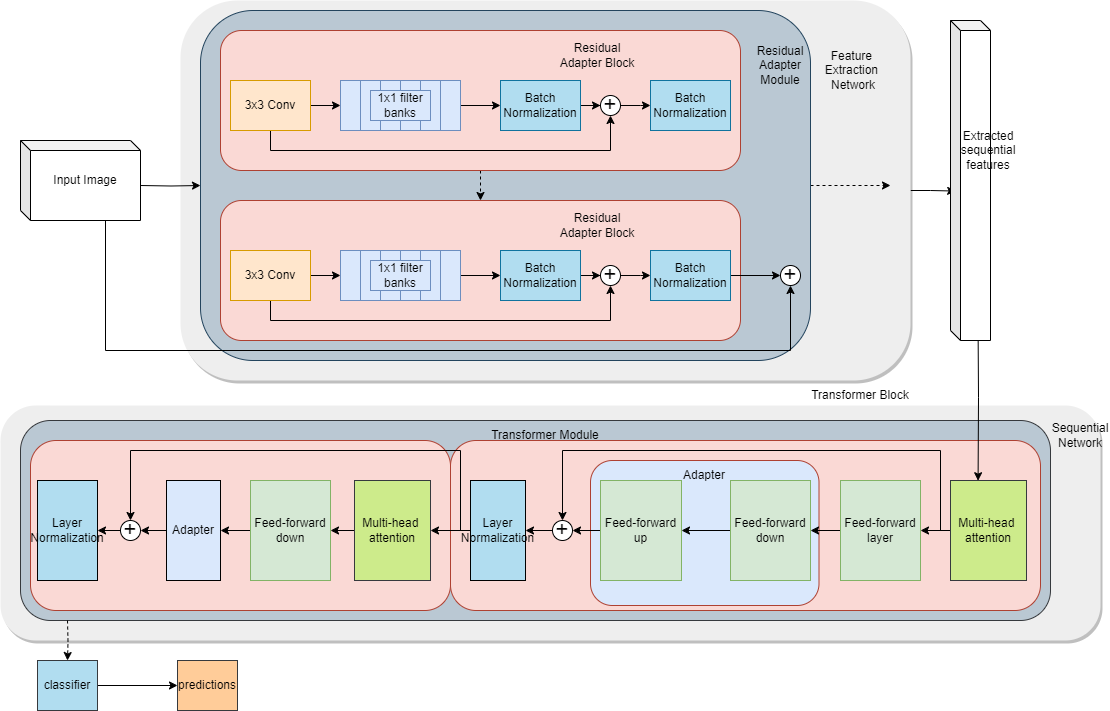}

}

\caption{\label{fig-schema}The proposed model's architecture. It
primarily comprises a feature extraction network and a sequential
network, augmented with adapter modules. This architecture is
schematically depicted with solid lines representing data flow and
dashed lines signifying the repetition of identical modules. Such a
configuration facilitates a modular and scalable design.}

\end{figure}

In the context of adapter tuning, the process is meticulously selective,
concentrating solely on the parameters of the adapters, normalization
layers, and the final classification layer, fostering a disentangled
form of learning. The network as a whole is subject to an end-to-end
training regimen. The model's architectural design, as depicted in
Figure~\ref{fig-schema}, exemplifies the cohesive interplay between
various innovative components designed to optimize character prediction
accuracy from input images.

\textbf{The feature extraction network.} Let's consider an input image
\(\mathcal{X} \in \mathbb{R}^{H \times W \times C}\), where \(H\),
\(W\), and \(C\) represent the height, width, and number of channels of
the image, respectively. The feature extraction network, modeled on the
principles of a Resnet-like convolutional neural network, operates on
\(\mathcal{X}\) to extract salient features. Denoted as
\(f_{\theta,\phi}(\mathcal{X})\), this network encompasses two types of
parameters: domain-agnostic parameters \(\theta\), which are common
across various domains, and domain-specific parameters \(\phi\), which
are unique to specific application areas. The network's architecture is
composed of a series of stacked residual modules, each containing one or
more residual blocks. These blocks are the fundamental units that
process the input image. Specifically, the \(i\)-th residual block in
the \(j\)-th module, \(g_{\theta_i,\phi_i}^{(j)}(\mathcal{X})\), can be
expressed as:
\(g_{\theta_i,\phi_i}^{(j)}(\mathcal{X}) = a((I + \alpha_i)(\omega_i*(\mathcal{X}))))\),
where \(i = 1,2,\dots\); \(a\) denotes the activation function; \(I\) is
the identity skip connection, facilitating gradient flow during
training; \(\omega_i\) are the domain-agnostic weights; \(\alpha_i\) are
the domain-specific weights; and \(*\) symbolizes the convolution
operation. For the sake of simplicity, normalization layers, typically
essential for stabilizing training, are not included in this formula.
Each residual module, represented as \(G_j(\mathcal{X})\), is the
functional composition of its constituent residual blocks:
\(G_j(\mathcal{X}) = g_1 \circ g_2 \circ \dots \circ g_n(\mathcal{X}),\)
with \(n\) indicating the number of residual blocks in the \(j\)-th
module and \(\circ\) symbolizing the composition of functions. Thus, the
feature extraction network can be holistically represented as:
\(f_{\theta,\phi}(\mathcal{X}) = G_1 \circ G_2 \circ \dots \circ G_m(\mathcal{X}),\)
where \(G_j\) denotes the \(j\)-th residual module and \(m\) the total
number of modules. The output of this network is
\(\mathcal{F} \in \mathbb{R}^{1 \times W' \times C'}\), where \(W'\) and
\(C'\) are the width and number of channels of the feature map,
respectively. Notably, the height is deliberately reduced to 1 for
seamless integration into subsequent sequential models.

\textbf{The sequential network}, as a core component of our model, is a
transformer-based architecture integrated with specialized bottleneck
adapter modules. This network takes the features extracted by the
feature extraction network as input and outputs the final predictions.
At the heart of this network are the bottleneck adapters, each
comprising three distinct layers: a linear down-projection layer, a
non-linearity layer, and a linear up-projection layer. These layers work
in tandem to refine the feature representation: the down-projection
layer condenses the input features from the transformer layer into a
lower-dimensional space, represented mathematically as
\(W_d\mathcal{X}\). Here, \(W_d\) denotes the weights of the
down-projection layer. After down-projection, a non-linear function,
symbolized by \(a\), is applied to these features. This step introduces
non-linearity to the model, enhancing its capacity to capture complex
patterns. Functions like ReLU or tanh are typically used for this
purpose. The transformed features are then projected back to their
original dimension through the up-projection layer, with weights
represented by \(W_u\). The adapter module can thus be denoted as
\(h_{w_d,w_u}(\mathcal{X}) = W_u a(W_d\mathcal{X})\). The
down-projection layer and up-projection layer are typically much smaller
than the transformer layer itself, which makes the adapter layers much
faster to train and allows them to be added to a larger number of layers
in the transformer model. These adapters are strategically placed after
the attention or feed-forward layers in the transformer. This placement
is critical as these layers are instrumental in learning complex
representations. By integrating adapters here, we can fine-tune the
model more effectively for specific tasks such as natural language
understanding, machine translation, and text summarization. This
approach has been corroborated by various studies (C. C. Liu et al.
2023; Mao et al. 2022; Rohanian et al. 2023; Hu et al. 2023; Houlsby et
al. 2019). The final stage involves the application of a softmax
function to convert the output into a probability distribution over the
output classes. The output \(\mathcal{Y}\), denoted as
\(\mathcal{Y} \in \mathbb{R}^{1 \times W' \times C'}\), reflects the
width \(W'\) and the number of classes \(C'\). To compute the loss,
\(\mathcal{Y}\) is fed into the Connectionist Temporal Classification
(CTC) loss function (Graves et al. 2006). This loss function is
particularly effective for sequence-to-sequence problems, providing a
robust framework for model training and optimization.

\textbf{Training the Adapter Modules.} The backbone of the network,
namely the feature extraction network and the sequential network, is
initially trained on a large dataset, with the exclusion of the adapter
modules. In general, the efficacy of the backbone in extracting useful
features correlates strongly with the diversity and size of the dataset.
Exclusively training on a limited dataset inside a singular domain has
the potential to result in overfitting, hence hindering the model's
ability to effectively generalize across other domains. The following
step of training involves incorporating a new task into the model and
subsequently optimizing the model's performance on it. The new task may
encompass a novel domain such as a different language, a distinct font,
or a new environmental configuration. The existing model has the
capability to be modified and applied to the new task while retaining
knowledge of the past tasks. The prevention of forgetting is
accomplished by freezing the weights of the backbone and solely updating
the weights of the corresponding adapter modules during the second phase
of training. The adapter modules can be conceptualized as a collection
of task-specific modules that are incorporated to enhance the efficiency
of feature extraction for the novel job. During the process of
backpropagation, the data originating from the new domain is exclusively
directed through its corresponding adapter module, while the remaining
adapter modules remain unaffected. Therefore, the performance of the
model on different domains is unaffected. Consequently, the model can
accurately identify characters from several domains without requiring
the training of separate models for each domain. By freezing the weights
of the backbone, the training process benefits from a significant
reduction in the number of parameters to be optimized. This leads to a
faster training period and mitigates the potential problem of
overfitting. The utilization of information from the backbone by the
adapters, which have been trained on a substantial dataset to extract
the most valuable characteristics, leads to a reduction in the needed
data size and number of training epochs for the adapter. Hence, our
model exhibits a high degree of adaptability in character recognition
across many domains, resulting in efficient utilization of training time
and resources.

\hypertarget{sec-experiments}{%
\section{Experiments}\label{sec-experiments}}

\textbf{Datasets.} We utilize the
\href{https://github.com/bupt-ai-cz/Meta-SelfLearning}{Meta
Self-Learning for Multi-Source Domain Adaptation: A Benchmark} dataset,
as documented by Qiu, Zhu, and Zhou (2021), to validate the efficacy of
our proposed model. This dataset, which is publicly available, comprises
various Chinese text images curated specifically for multi-source domain
adaptation studies. The selection of a Chinese dataset is motivated by
the linguistic complexity of Chinese, with its character set exceeding
10,000 unique glyphs (11,376 in our study), as opposed to the mere 35
characters found in traditional English-digit datasets. Such complexity
presents a greater challenge in character recognition, potentially
enhancing the network's capacity for general feature learning and
subsequent domain adaptability.

The dataset is categorized into five distinct text image types:
handwritten, street scene, document, synthetic, and car license plates.
For the initial experiment, the document category---consisting of
1,614,955 training and 179,345 test images---is employed to train the
network's backbone. These images, which feature a uniform font and clean
background, serve as an ideal starting point. Subsequently, we evaluate
the model's domain adaptation capability using two additional
categories: car license plates and synthetic images. The car license
plate domain, containing 187,136 training and 20,792 test images, is
considered less challenging due to the uniformity of font, clean
background, and limited character set. Conversely, the synthetic domain,
with its diverse fonts, noisy backgrounds, and non-meaningful synthetic
sentences, creates a more strenuous test environment. This category
includes 999,558 training and 111,062 test images. Each class in the
dataset represents a distinct Chinese character, totaling 11,376
classes. The images are preprocessed to a standard size of 32x128 pixels
and normalized to ensure a mean of zero and unit variance, facilitating
consistent input for the network.

\textbf{Implementation Details.} Our feature extraction network employs
a relatively shallow architecture due to the size of our training data.
It comprises 4 residual modules, each containing 2 residual blocks. The
sequential network utilizes a multi-head attention mechanism with 8
heads and incorporates positional encoding to capture sequence
information. Network feedforward layers consist of 128 hidden units
each.

For the training regimen, we employ a batch size of 256 and the Adam
optimizer with an initial learning rate of \(1 \times 10^{-5}\). The
learning rate is subjected to a decay by a factor of 0.1 every five
epochs. The first two epochs function as a warmup phase for the learning
rate. We train the backbone for a total of 20 epochs under two distinct
experimental scenarios to evaluate the adaptability of the model:

\begin{enumerate}
\def\labelenumi{\arabic{enumi}.}
\tightlist
\item
  In the first scenario, the backbone is exclusively trained using the
  document dataset to test the adapters' generalization to unseen
  datasets.
\item
  In the second scenario, the backbone training also includes data from
  the car license and synthetic datasets to explore how well the
  adapters can optimize performance within specific domains.
\end{enumerate}

Following the backbone training, we train the adapter models on the car
license dataset or the synthetic dataset for 20 additional epochs. We
compare the outcomes of exclusively training the adapter modules
(adapter method) versus jointly training the adapter modules and the
backbone (finetuning method).

Performance is assessed using three metrics: character accuracy, word
accuracy, and recall. Character accuracy (precision) is the fraction of
correctly identified characters in the OCR output relative to the ground
truth. Word accuracy, effectively image-level accuracy due to the
absence of Chinese word segmentation, measures correct word recognition.
Recall is defined as \(Recall = \frac{TP}{TP + FN}\), where \(TP\) is
the count of accurately identified characters or words, and \(FN\)
represents those that the OCR system failed to identify. As an OCR
engine tuned for high recall may inadvertently reduce precision,
achieving a balance between these metrics is vital. This consideration
is particularly crucial in fields that demand comprehensive data
extraction, such as legal and medical document processing. Our model is
developed in PyTorch and trained on a suite of eight NVIDIA Tesla K80
GPUs. The source code is available at
https://github.com/Jiayou-Chao/Multi-domain-OCR.

\textbf{Training Backone Results.} The details pertaining to the
backbone models can be found in Table~\ref{tbl-backbone-results}. In the
1st experiment, the backbone is solely trained using the document
dataset. The evaluation is conducted on all three datasets, namely
document, car license, and synthetic. The backbone model demonstrates a
character accuracy rate of 94.93\%, a word accuracy rate of 65.15\%, and
a recall rate of 94.91\% when evaluated on the document dataset. The
metrics exhibit values in proximity to zero when the model undergoes
evaluation on the car license or the synthetic dataset. This observation
indicates that the current backbone model lacks the capacity to
generalize effectively to unfamiliar domains. Consequently, we will
employ this model as a means to evaluate the efficacy of adapters in
facilitating generalization to unseen domains. In the 2nd experiment,
the backbone undergoes training and subsequent evaluation on all three
datasets. On the document dataset, the backbone model exhibits a
character accuracy of 99.29\%, a word accuracy of 93.23\%, and a recall
rate of 99.26\%. The current backbone demonstrates superior performance
across all metrics in comparison to the previous experiment, which
solely utilized the document dataset for training. The performance
metrics of the backbone model on the car license dataset are as follows:
99.80\% character accuracy, 98.52\% word accuracy, and 99.79\% recall.
The performance metrics of the backbone model on the synthetic dataset
are as follows: 92.17\% character accuracy, 64.02\% word accuracy, and
92.13\% recall. This observation indicates that there is potential for
improvement in the model's performance on the synthetic dataset,
particularly in terms of word accuracy. The performance exhibited in
this instance is highly commendable; however, it is important to
acknowledge that the second backbone has been trained on a significantly
larger dataset, thereby necessitating a substantial investment of time
and computational resources. The backbone model is subsequently employed
to train the adapter models in order to evaluate their potential for
improving performance within a particular domain, while maintaining the
same level of performance across other domains.

\hypertarget{tbl-backbone-results}{}
\global\setlength{\Oldarrayrulewidth}{\arrayrulewidth}

\global\setlength{\Oldtabcolsep}{\tabcolsep}

\setlength{\tabcolsep}{0pt}

\renewcommand*{\arraystretch}{1.5}

\providecommand{\ascline}[3]{\noalign{\global\arrayrulewidth #1}\arrayrulecolor[HTML]{#2}\cline{#3}}

\begin{longtable}[c]{|p{1.57in}|p{1.63in}|p{1.32in}|p{0.80in}}
\caption{\label{tbl-backbone-results}The information of the backbone models. In the 1st experiment, the
backbone is trained only on the document dataset. The model performs
poorly on unseen datasets like the car license dataset and the synthetic
dataset. In the 2nd experiment, the backbone is trained on all of the
three datasets (document, car license, and synthetic). Three metrics,
character accuracy, word accuracy and recall, are used to evaluate the
backbone on all three datasets. }\tabularnewline

\ascline{1.5pt}{666666}{1-4}

\multicolumn{1}{>{\raggedright}m{\dimexpr 1.57in+0\tabcolsep}}{\textcolor[HTML]{000000}{\fontsize{11}{11}\selectfont{Evaluation\ Dataset}}} & \multicolumn{1}{>{\raggedright}m{\dimexpr 1.63in+0\tabcolsep}}{\textcolor[HTML]{000000}{\fontsize{11}{11}\selectfont{Character\ Accuracy}}} & \multicolumn{1}{>{\raggedright}m{\dimexpr 1.32in+0\tabcolsep}}{\textcolor[HTML]{000000}{\fontsize{11}{11}\selectfont{Word\ Accuracy}}} & \multicolumn{1}{>{\raggedright}m{\dimexpr 0.8in+0\tabcolsep}}{\textcolor[HTML]{000000}{\fontsize{11}{11}\selectfont{Recall}}} \\

\ascline{1.5pt}{666666}{1-4}\endfirsthead 

\ascline{1.5pt}{666666}{1-4}

\multicolumn{1}{>{\raggedright}m{\dimexpr 1.57in+0\tabcolsep}}{\textcolor[HTML]{000000}{\fontsize{11}{11}\selectfont{Evaluation\ Dataset}}} & \multicolumn{1}{>{\raggedright}m{\dimexpr 1.63in+0\tabcolsep}}{\textcolor[HTML]{000000}{\fontsize{11}{11}\selectfont{Character\ Accuracy}}} & \multicolumn{1}{>{\raggedright}m{\dimexpr 1.32in+0\tabcolsep}}{\textcolor[HTML]{000000}{\fontsize{11}{11}\selectfont{Word\ Accuracy}}} & \multicolumn{1}{>{\raggedright}m{\dimexpr 0.8in+0\tabcolsep}}{\textcolor[HTML]{000000}{\fontsize{11}{11}\selectfont{Recall}}} \\

\ascline{1.5pt}{666666}{1-4}\endhead

\multicolumn{4}{>{\raggedright}m{\dimexpr 5.32in+6\tabcolsep}}{\textcolor[HTML]{000000}{\fontsize{11}{11}\selectfont{\textbf{Experiment}}}\textcolor[HTML]{000000}{\fontsize{11}{11}\selectfont{\textbf{:\ }}}\textcolor[HTML]{000000}{\fontsize{11}{11}\selectfont{\textbf{1}}}} \\

\ascline{1pt}{666666}{1-4}

\multicolumn{1}{>{\raggedright}m{\dimexpr 1.57in+0\tabcolsep}}{\textcolor[HTML]{000000}{\fontsize{11}{11}\selectfont{document}}} & \multicolumn{1}{>{\raggedright}m{\dimexpr 1.63in+0\tabcolsep}}{\textcolor[HTML]{000000}{\fontsize{11}{11}\selectfont{94.93\%}}} & \multicolumn{1}{>{\raggedright}m{\dimexpr 1.32in+0\tabcolsep}}{\textcolor[HTML]{000000}{\fontsize{11}{11}\selectfont{65.15\%}}} & \multicolumn{1}{>{\raggedright}m{\dimexpr 0.8in+0\tabcolsep}}{\textcolor[HTML]{000000}{\fontsize{11}{11}\selectfont{94.91\%}}} \\

\multicolumn{1}{>{\raggedright}m{\dimexpr 1.57in+0\tabcolsep}}{\textcolor[HTML]{000000}{\fontsize{11}{11}\selectfont{car\ license}}} & \multicolumn{1}{>{\raggedright}m{\dimexpr 1.63in+0\tabcolsep}}{\textcolor[HTML]{000000}{\fontsize{11}{11}\selectfont{2.44\%}}} & \multicolumn{1}{>{\raggedright}m{\dimexpr 1.32in+0\tabcolsep}}{\textcolor[HTML]{000000}{\fontsize{11}{11}\selectfont{0.00\%}}} & \multicolumn{1}{>{\raggedright}m{\dimexpr 0.8in+0\tabcolsep}}{\textcolor[HTML]{000000}{\fontsize{11}{11}\selectfont{1.84\%}}} \\

\multicolumn{1}{>{\raggedright}m{\dimexpr 1.57in+0\tabcolsep}}{\textcolor[HTML]{000000}{\fontsize{11}{11}\selectfont{synthetic}}} & \multicolumn{1}{>{\raggedright}m{\dimexpr 1.63in+0\tabcolsep}}{\textcolor[HTML]{000000}{\fontsize{11}{11}\selectfont{0.43\%}}} & \multicolumn{1}{>{\raggedright}m{\dimexpr 1.32in+0\tabcolsep}}{\textcolor[HTML]{000000}{\fontsize{11}{11}\selectfont{0.00\%}}} & \multicolumn{1}{>{\raggedright}m{\dimexpr 0.8in+0\tabcolsep}}{\textcolor[HTML]{000000}{\fontsize{11}{11}\selectfont{0.55\%}}} \\

\ascline{1pt}{666666}{1-4}

\multicolumn{4}{>{\raggedright}m{\dimexpr 5.32in+6\tabcolsep}}{\textcolor[HTML]{000000}{\fontsize{11}{11}\selectfont{\textbf{Experiment}}}\textcolor[HTML]{000000}{\fontsize{11}{11}\selectfont{\textbf{:\ }}}\textcolor[HTML]{000000}{\fontsize{11}{11}\selectfont{\textbf{2}}}} \\

\ascline{1pt}{666666}{1-4}

\multicolumn{1}{>{\raggedright}m{\dimexpr 1.57in+0\tabcolsep}}{\textcolor[HTML]{000000}{\fontsize{11}{11}\selectfont{document}}} & \multicolumn{1}{>{\raggedright}m{\dimexpr 1.63in+0\tabcolsep}}{\textcolor[HTML]{000000}{\fontsize{11}{11}\selectfont{99.29\%}}} & \multicolumn{1}{>{\raggedright}m{\dimexpr 1.32in+0\tabcolsep}}{\textcolor[HTML]{000000}{\fontsize{11}{11}\selectfont{93.32\%}}} & \multicolumn{1}{>{\raggedright}m{\dimexpr 0.8in+0\tabcolsep}}{\textcolor[HTML]{000000}{\fontsize{11}{11}\selectfont{99.26\%}}} \\

\multicolumn{1}{>{\raggedright}m{\dimexpr 1.57in+0\tabcolsep}}{\textcolor[HTML]{000000}{\fontsize{11}{11}\selectfont{car\ license}}} & \multicolumn{1}{>{\raggedright}m{\dimexpr 1.63in+0\tabcolsep}}{\textcolor[HTML]{000000}{\fontsize{11}{11}\selectfont{99.80\%}}} & \multicolumn{1}{>{\raggedright}m{\dimexpr 1.32in+0\tabcolsep}}{\textcolor[HTML]{000000}{\fontsize{11}{11}\selectfont{98.52\%}}} & \multicolumn{1}{>{\raggedright}m{\dimexpr 0.8in+0\tabcolsep}}{\textcolor[HTML]{000000}{\fontsize{11}{11}\selectfont{99.79\%}}} \\

\multicolumn{1}{>{\raggedright}m{\dimexpr 1.57in+0\tabcolsep}}{\textcolor[HTML]{000000}{\fontsize{11}{11}\selectfont{synthetic}}} & \multicolumn{1}{>{\raggedright}m{\dimexpr 1.63in+0\tabcolsep}}{\textcolor[HTML]{000000}{\fontsize{11}{11}\selectfont{92.17\%}}} & \multicolumn{1}{>{\raggedright}m{\dimexpr 1.32in+0\tabcolsep}}{\textcolor[HTML]{000000}{\fontsize{11}{11}\selectfont{64.02\%}}} & \multicolumn{1}{>{\raggedright}m{\dimexpr 0.8in+0\tabcolsep}}{\textcolor[HTML]{000000}{\fontsize{11}{11}\selectfont{92.13\%}}} \\

\ascline{1.5pt}{666666}{1-4}

\end{longtable}

\arrayrulecolor[HTML]{000000}

\global\setlength{\arrayrulewidth}{\Oldarrayrulewidth}

\global\setlength{\tabcolsep}{\Oldtabcolsep}

\renewcommand*{\arraystretch}{1}

\textbf{Training Adapter Results.} The results of training only adapters
in two differenet scenarios are shown in Table~\ref{tbl-adapter-results}
and the results are compared with the finetuning method which updates
all parameters in comparison to updating only adapters. When using the
finetuning method, there are a total of 21,636,658 trainable parameters
(the parameters of the irrelevant domains are excluded when
calculating). In contrast, there are only 7,590,930 trainable parameters
when using the adapter method, which is a marked reduction of 64.93\%.
Delving into the specifics, Experiment 1 unveils an enhancement in
character accuracy from a meager 2.44\% to an impressive 99.64\% using
adapters and a slightly higher 99.83\% with finetuning on the car
license dataset. This is a dataset that is considered relatively
straightforward due to its simplicity, hinting that the adapter method
can indeed parallel the performance of full finetuning on simple tasks.
Moreover, this method has the added advantage of retaining prior
knowledge without the risk of domain forgetting---a common hurdle in
finetuning. When examining word accuracy, the adapter method even
outperforms finetuning by achieving 99.63\% compared to 98.81\%.
However, the recall rate with the adapter method at 97.50\%
underperforms the finetuning method at 99.83\%. In the context of the
synthetic dataset, which poses a greater challenge due to its inclusion
of unseen characters and nonsensical sentences, the adapter method
maintains its competence. Character accuracy ascends from a paltry
0.43\% to 96.13\% and recall from 0.55\% to 96.04\% for the adapter
method, nearing the finetuning outcomes of 98.94\% and 93.91\%,
respectively. Although the word accuracy for adapter lags at 79.81\%
versus the finetuning's 98.89\%, the marginal disparity highlights a
limitation in the backbone's generalization capabilities to unfamiliar
domains. The evidence, as outlined here, suggests that adapters offer a
promising alternative to full model finetuning, particularly in
scenarios where computational efficiency and memory preservation are
paramount. Despite the occasional dips in performance on more complex
datasets, the adapter method's impressive recall and character accuracy
signify its potential as a viable strategy for domain-specific neural
network training.

Experiment 2 offers an extension of the inquiry into the efficacy of
adapter training versus full model finetuning, this time with the
backbone exposed to a more diverse training set. Indeed, the performance
on the car license datasets exhibited minimal variance, which can be
attributed to the already optimized state of the backbone for this
particular domain, evidencing a saturation point in learning. On the
synthetic dataset, which serves as the benchmark for complexity within
this study, both methods demonstrated significant improvements in
character and word accuracies, achieving near-parity. The adapter method
enhanced character accuracy from 92.17\% to 98.48\%, while finetuning
edged slightly ahead with a rise to 98.84\%. Moreover, word accuracy for
adapters soared from 64.02\% to a remarkable 98.44\%, closely shadowing
the finetuning result of 98.89\%. Interestingly, recall rates exhibited
a different trend, remaining relatively stable for the adapter method
with a slight decline from 92.13\% to 91.56\%, whereas finetuning saw a
modest increase to 93.39\%.

These outcomes are particularly noteworthy in that they illustrate the
adapter method's robustness when the backbone network is enriched with
diverse training data. In essence, the adapter-equipped model nearly
matches the finetuning method in performance, even when confronted with
complex datasets. Moreover, it underscores the adapter method's
proficiency in domain-specific enhancement without compromising the
existing knowledge encoded in the backbone---a notable advantage over
training all domains concurrently on the backbone. Thus, Experiment 2
reinforces the conclusion drawn from the initial experiment: the adapter
method not only boasts fewer parameters and reduced risk of catastrophic
forgetting but also exhibits the potential to deliver performance
comparable to finetuning, even under the pressures of dataset
complexity. Given the additional benefits of selective domain
enhancement, the adapter method emerges as a compelling choice for
efficient and effective neural network training, especially when
considering the computational and memory constraints often encountered
in real-world applications.

\hypertarget{tbl-adapter-results}{}
\global\setlength{\Oldarrayrulewidth}{\arrayrulewidth}

\global\setlength{\Oldtabcolsep}{\tabcolsep}

\setlength{\tabcolsep}{0pt}

\renewcommand*{\arraystretch}{1.5}

\providecommand{\ascline}[3]{\noalign{\global\arrayrulewidth #1}\arrayrulecolor[HTML]{#2}\cline{#3}}

\begin{longtable}[c]{|p{1.00in}|p{1.57in}|p{1.25in}|p{0.95in}|p{0.80in}|p{1.41in}}
\caption{\label{tbl-adapter-results}The training and comparison results of the adapter. The adapter method
refers to the training of adapter modules while keeping the backbone
frozen. The finetuning approach involves training the adapter modules
and the backbone simultaneously. In addition to evaluating character
accuracy, word accuracy, and recall, we also include the number of
trainable parameters in the last table column. Consequently, the adapter
method shows comparable performance to the finetuning method on a simple
new domain, while its effectiveness is constrained by the underlying
backbone on a more intricate new domain. The adapter method does not
have the risk of forgetting the previous domain. }\tabularnewline

\ascline{1.5pt}{666666}{1-6}

\multicolumn{1}{>{\raggedright}m{\dimexpr 1in+0\tabcolsep}}{\textcolor[HTML]{000000}{\fontsize{11}{11}\selectfont{Method}}} & \multicolumn{1}{>{\raggedright}m{\dimexpr 1.57in+0\tabcolsep}}{\textcolor[HTML]{000000}{\fontsize{11}{11}\selectfont{Evaluation\ Dataset}}} & \multicolumn{1}{>{\raggedright}m{\dimexpr 1.25in+0\tabcolsep}}{\textcolor[HTML]{000000}{\fontsize{11}{11}\selectfont{Character\ Acc}}} & \multicolumn{1}{>{\raggedright}m{\dimexpr 0.95in+0\tabcolsep}}{\textcolor[HTML]{000000}{\fontsize{11}{11}\selectfont{Word\ Acc}}} & \multicolumn{1}{>{\raggedright}m{\dimexpr 0.8in+0\tabcolsep}}{\textcolor[HTML]{000000}{\fontsize{11}{11}\selectfont{Recall}}} & \multicolumn{1}{>{\raggedleft}m{\dimexpr 1.41in+0\tabcolsep}}{\textcolor[HTML]{000000}{\fontsize{11}{11}\selectfont{Trainable\ Param}}} \\

\ascline{1.5pt}{666666}{1-6}\endfirsthead 

\ascline{1.5pt}{666666}{1-6}

\multicolumn{1}{>{\raggedright}m{\dimexpr 1in+0\tabcolsep}}{\textcolor[HTML]{000000}{\fontsize{11}{11}\selectfont{Method}}} & \multicolumn{1}{>{\raggedright}m{\dimexpr 1.57in+0\tabcolsep}}{\textcolor[HTML]{000000}{\fontsize{11}{11}\selectfont{Evaluation\ Dataset}}} & \multicolumn{1}{>{\raggedright}m{\dimexpr 1.25in+0\tabcolsep}}{\textcolor[HTML]{000000}{\fontsize{11}{11}\selectfont{Character\ Acc}}} & \multicolumn{1}{>{\raggedright}m{\dimexpr 0.95in+0\tabcolsep}}{\textcolor[HTML]{000000}{\fontsize{11}{11}\selectfont{Word\ Acc}}} & \multicolumn{1}{>{\raggedright}m{\dimexpr 0.8in+0\tabcolsep}}{\textcolor[HTML]{000000}{\fontsize{11}{11}\selectfont{Recall}}} & \multicolumn{1}{>{\raggedleft}m{\dimexpr 1.41in+0\tabcolsep}}{\textcolor[HTML]{000000}{\fontsize{11}{11}\selectfont{Trainable\ Param}}} \\

\ascline{1.5pt}{666666}{1-6}\endhead

\multicolumn{6}{>{\raggedright}m{\dimexpr 6.98in+10\tabcolsep}}{\textcolor[HTML]{000000}{\fontsize{11}{11}\selectfont{\textbf{Experiment}}}\textcolor[HTML]{000000}{\fontsize{11}{11}\selectfont{\textbf{:\ }}}\textcolor[HTML]{000000}{\fontsize{11}{11}\selectfont{\textbf{1}}}} \\

\ascline{1pt}{666666}{1-6}

\multicolumn{1}{>{\raggedright}m{\dimexpr 1in+0\tabcolsep}}{\textcolor[HTML]{000000}{\fontsize{11}{11}\selectfont{Adapter}}} & \multicolumn{1}{>{\raggedright}m{\dimexpr 1.57in+0\tabcolsep}}{\textcolor[HTML]{000000}{\fontsize{11}{11}\selectfont{car\ license}}} & \multicolumn{1}{>{\raggedright}m{\dimexpr 1.25in+0\tabcolsep}}{\textcolor[HTML]{000000}{\fontsize{11}{11}\selectfont{99.64\%}}} & \multicolumn{1}{>{\raggedright}m{\dimexpr 0.95in+0\tabcolsep}}{\textcolor[HTML]{000000}{\fontsize{11}{11}\selectfont{99.63\%}}} & \multicolumn{1}{>{\raggedright}m{\dimexpr 0.8in+0\tabcolsep}}{\textcolor[HTML]{000000}{\fontsize{11}{11}\selectfont{97.50\%}}} & \multicolumn{1}{>{\raggedleft}m{\dimexpr 1.41in+0\tabcolsep}}{\textcolor[HTML]{000000}{\fontsize{11}{11}\selectfont{7,590,930}}} \\

\multicolumn{1}{>{\raggedright}m{\dimexpr 1in+0\tabcolsep}}{\textcolor[HTML]{000000}{\fontsize{11}{11}\selectfont{Finetuning}}} & \multicolumn{1}{>{\raggedright}m{\dimexpr 1.57in+0\tabcolsep}}{\textcolor[HTML]{000000}{\fontsize{11}{11}\selectfont{car\ license}}} & \multicolumn{1}{>{\raggedright}m{\dimexpr 1.25in+0\tabcolsep}}{\textcolor[HTML]{000000}{\fontsize{11}{11}\selectfont{99.83\%}}} & \multicolumn{1}{>{\raggedright}m{\dimexpr 0.95in+0\tabcolsep}}{\textcolor[HTML]{000000}{\fontsize{11}{11}\selectfont{98.81\%}}} & \multicolumn{1}{>{\raggedright}m{\dimexpr 0.8in+0\tabcolsep}}{\textcolor[HTML]{000000}{\fontsize{11}{11}\selectfont{99.83\%}}} & \multicolumn{1}{>{\raggedleft}m{\dimexpr 1.41in+0\tabcolsep}}{\textcolor[HTML]{000000}{\fontsize{11}{11}\selectfont{21,636,658}}} \\

\multicolumn{1}{>{\raggedright}m{\dimexpr 1in+0\tabcolsep}}{\textcolor[HTML]{000000}{\fontsize{11}{11}\selectfont{Adapter}}} & \multicolumn{1}{>{\raggedright}m{\dimexpr 1.57in+0\tabcolsep}}{\textcolor[HTML]{000000}{\fontsize{11}{11}\selectfont{synthetic}}} & \multicolumn{1}{>{\raggedright}m{\dimexpr 1.25in+0\tabcolsep}}{\textcolor[HTML]{000000}{\fontsize{11}{11}\selectfont{96.13\%}}} & \multicolumn{1}{>{\raggedright}m{\dimexpr 0.95in+0\tabcolsep}}{\textcolor[HTML]{000000}{\fontsize{11}{11}\selectfont{79.81\%}}} & \multicolumn{1}{>{\raggedright}m{\dimexpr 0.8in+0\tabcolsep}}{\textcolor[HTML]{000000}{\fontsize{11}{11}\selectfont{96.04\%}}} & \multicolumn{1}{>{\raggedleft}m{\dimexpr 1.41in+0\tabcolsep}}{\textcolor[HTML]{000000}{\fontsize{11}{11}\selectfont{7,590,930}}} \\

\multicolumn{1}{>{\raggedright}m{\dimexpr 1in+0\tabcolsep}}{\textcolor[HTML]{000000}{\fontsize{11}{11}\selectfont{Finetuning}}} & \multicolumn{1}{>{\raggedright}m{\dimexpr 1.57in+0\tabcolsep}}{\textcolor[HTML]{000000}{\fontsize{11}{11}\selectfont{synthetic}}} & \multicolumn{1}{>{\raggedright}m{\dimexpr 1.25in+0\tabcolsep}}{\textcolor[HTML]{000000}{\fontsize{11}{11}\selectfont{98.94\%}}} & \multicolumn{1}{>{\raggedright}m{\dimexpr 0.95in+0\tabcolsep}}{\textcolor[HTML]{000000}{\fontsize{11}{11}\selectfont{98.89\%}}} & \multicolumn{1}{>{\raggedright}m{\dimexpr 0.8in+0\tabcolsep}}{\textcolor[HTML]{000000}{\fontsize{11}{11}\selectfont{93.91\%}}} & \multicolumn{1}{>{\raggedleft}m{\dimexpr 1.41in+0\tabcolsep}}{\textcolor[HTML]{000000}{\fontsize{11}{11}\selectfont{21,636,658}}} \\

\ascline{1pt}{666666}{1-6}

\multicolumn{6}{>{\raggedright}m{\dimexpr 6.98in+10\tabcolsep}}{\textcolor[HTML]{000000}{\fontsize{11}{11}\selectfont{\textbf{Experiment}}}\textcolor[HTML]{000000}{\fontsize{11}{11}\selectfont{\textbf{:\ }}}\textcolor[HTML]{000000}{\fontsize{11}{11}\selectfont{\textbf{2}}}} \\

\ascline{1pt}{666666}{1-6}

\multicolumn{1}{>{\raggedright}m{\dimexpr 1in+0\tabcolsep}}{\textcolor[HTML]{000000}{\fontsize{11}{11}\selectfont{Adapter}}} & \multicolumn{1}{>{\raggedright}m{\dimexpr 1.57in+0\tabcolsep}}{\textcolor[HTML]{000000}{\fontsize{11}{11}\selectfont{car\ license}}} & \multicolumn{1}{>{\raggedright}m{\dimexpr 1.25in+0\tabcolsep}}{\textcolor[HTML]{000000}{\fontsize{11}{11}\selectfont{99.64\%}}} & \multicolumn{1}{>{\raggedright}m{\dimexpr 0.95in+0\tabcolsep}}{\textcolor[HTML]{000000}{\fontsize{11}{11}\selectfont{99.63\%}}} & \multicolumn{1}{>{\raggedright}m{\dimexpr 0.8in+0\tabcolsep}}{\textcolor[HTML]{000000}{\fontsize{11}{11}\selectfont{97.50\%}}} & \multicolumn{1}{>{\raggedleft}m{\dimexpr 1.41in+0\tabcolsep}}{\textcolor[HTML]{000000}{\fontsize{11}{11}\selectfont{7,590,930}}} \\

\multicolumn{1}{>{\raggedright}m{\dimexpr 1in+0\tabcolsep}}{\textcolor[HTML]{000000}{\fontsize{11}{11}\selectfont{Finetuning}}} & \multicolumn{1}{>{\raggedright}m{\dimexpr 1.57in+0\tabcolsep}}{\textcolor[HTML]{000000}{\fontsize{11}{11}\selectfont{car\ license}}} & \multicolumn{1}{>{\raggedright}m{\dimexpr 1.25in+0\tabcolsep}}{\textcolor[HTML]{000000}{\fontsize{11}{11}\selectfont{99.97\%}}} & \multicolumn{1}{>{\raggedright}m{\dimexpr 0.95in+0\tabcolsep}}{\textcolor[HTML]{000000}{\fontsize{11}{11}\selectfont{99.97\%}}} & \multicolumn{1}{>{\raggedright}m{\dimexpr 0.8in+0\tabcolsep}}{\textcolor[HTML]{000000}{\fontsize{11}{11}\selectfont{99.97\%}}} & \multicolumn{1}{>{\raggedleft}m{\dimexpr 1.41in+0\tabcolsep}}{\textcolor[HTML]{000000}{\fontsize{11}{11}\selectfont{21,636,658}}} \\

\multicolumn{1}{>{\raggedright}m{\dimexpr 1in+0\tabcolsep}}{\textcolor[HTML]{000000}{\fontsize{11}{11}\selectfont{Adapter}}} & \multicolumn{1}{>{\raggedright}m{\dimexpr 1.57in+0\tabcolsep}}{\textcolor[HTML]{000000}{\fontsize{11}{11}\selectfont{synthetic}}} & \multicolumn{1}{>{\raggedright}m{\dimexpr 1.25in+0\tabcolsep}}{\textcolor[HTML]{000000}{\fontsize{11}{11}\selectfont{98.48\%}}} & \multicolumn{1}{>{\raggedright}m{\dimexpr 0.95in+0\tabcolsep}}{\textcolor[HTML]{000000}{\fontsize{11}{11}\selectfont{98.44\%}}} & \multicolumn{1}{>{\raggedright}m{\dimexpr 0.8in+0\tabcolsep}}{\textcolor[HTML]{000000}{\fontsize{11}{11}\selectfont{91.56\%}}} & \multicolumn{1}{>{\raggedleft}m{\dimexpr 1.41in+0\tabcolsep}}{\textcolor[HTML]{000000}{\fontsize{11}{11}\selectfont{7,590,930}}} \\

\multicolumn{1}{>{\raggedright}m{\dimexpr 1in+0\tabcolsep}}{\textcolor[HTML]{000000}{\fontsize{11}{11}\selectfont{Finetuning}}} & \multicolumn{1}{>{\raggedright}m{\dimexpr 1.57in+0\tabcolsep}}{\textcolor[HTML]{000000}{\fontsize{11}{11}\selectfont{synthetic}}} & \multicolumn{1}{>{\raggedright}m{\dimexpr 1.25in+0\tabcolsep}}{\textcolor[HTML]{000000}{\fontsize{11}{11}\selectfont{98.94\%}}} & \multicolumn{1}{>{\raggedright}m{\dimexpr 0.95in+0\tabcolsep}}{\textcolor[HTML]{000000}{\fontsize{11}{11}\selectfont{98.89\%}}} & \multicolumn{1}{>{\raggedright}m{\dimexpr 0.8in+0\tabcolsep}}{\textcolor[HTML]{000000}{\fontsize{11}{11}\selectfont{93.90\%}}} & \multicolumn{1}{>{\raggedleft}m{\dimexpr 1.41in+0\tabcolsep}}{\textcolor[HTML]{000000}{\fontsize{11}{11}\selectfont{21,636,658}}} \\

\ascline{1.5pt}{666666}{1-6}

\end{longtable}

\arrayrulecolor[HTML]{000000}

\global\setlength{\arrayrulewidth}{\Oldarrayrulewidth}

\global\setlength{\tabcolsep}{\Oldtabcolsep}

\renewcommand*{\arraystretch}{1}

\hypertarget{sec-conclusion}{%
\section{Conclusion}\label{sec-conclusion}}

In this paper, we introduce an innovative adapter network designed for
multi-source OCR and present its effectiveness over traditional domain
adaptation methods. The empirical evidence from the conducted
experiments indicates that the adapter network outstrips the traditional
methods. In comparison to the alternative strategy, which involves
domain-specific fine-tuning of the backbone model, the adapter network
shows an equivalent aptitude in performance. However, the standout
feature of the adapter network is its reduction in the quantity of
parameters that require training. This reduction is not trivial---it
significantly streamlines the process of adapting to new domains, a
vital factor in business environments that demand both quick
adaptability and the capacity to handle multiple domains simultaneously.

Notwithstanding these advantages, the study also reveals a shortcoming
of the adapter network when dealing with complex domains---as evidenced
by the results from the synthetic dataset. When the backbone of the
network is relatively weak, the ability of the adapter network to
achieve commendable accuracy in intricate domains is compromised. The
findings underscore the necessity of training a robust backbone model on
an extensive dataset. Such comprehensive training is imperative for the
model to discern and assimilate the essential characteristics inherent
to the entities within the domain, thereby enhancing the model's
capability to render high accuracy in demanding and complex domains.

In essence, the research posits that while the adapter network holds
promise for flexible and efficient domain adaptation, the strength of
the underlying model is a fundamental precept that governs the ultimate
performance and robustness in challenging domain-specific tasks.

\hypertarget{sec-references}{%
\section*{References}\label{sec-references}}
\addcontentsline{toc}{section}{References}

\hypertarget{refs}{}
\begin{CSLReferences}{1}{0}
\leavevmode\vadjust pre{\hypertarget{ref-bengio2014representation}{}}%
Bengio, Yoshua, Aaron Courville, and Pascal Vincent. 2014.
{``Representation Learning: A Review and New Perspectives.''}
\url{https://arxiv.org/abs/1206.5538}.

\leavevmode\vadjust pre{\hypertarget{ref-bhattacharjee2023vision}{}}%
Bhattacharjee, Deblina, Sabine Süsstrunk, and Mathieu Salzmann. 2023.
{``Vision Transformer Adapters for Generalizable Multitask Learning.''}
\url{https://arxiv.org/abs/2308.12372}.

\leavevmode\vadjust pre{\hypertarget{ref-cannings2019classification}{}}%
Cannings, Timothy I., Yingying Fan, and Richard J. Samworth. 2019.
{``Classification with Imperfect Training Labels.''}
\url{https://arxiv.org/abs/1805.11505}.

\leavevmode\vadjust pre{\hypertarget{ref-caruana1997multitask}{}}%
Caruana, Rich. 1997. {``Multitask Learning.''} \emph{Machine Learning}
28 (1): 41--75.

\leavevmode\vadjust pre{\hypertarget{ref-Fayek_Cavedon_Wu_2020}{}}%
Fayek, Haytham M., Lawrence Cavedon, and Hong Ren Wu. 2020.
{``Progressive Learning: A Deep Learning Framework for Continual
Learning.''} \emph{Neural Networks} 128 (August): 345--57.
\url{https://doi.org/10.1016/j.neunet.2020.05.011}.

\leavevmode\vadjust pre{\hypertarget{ref-french1999catastrophic}{}}%
French, Robert M. 1999. {``Catastrophic Forgetting in Connectionist
Networks.''} \emph{Trends in Cognitive Sciences} 3 (4): 128--35.

\leavevmode\vadjust pre{\hypertarget{ref-ganin2016domainadversarial}{}}%
Ganin, Yaroslav, Evgeniya Ustinova, Hana Ajakan, Pascal Germain, Hugo
Larochelle, François Laviolette, Mario Marchand, and Victor Lempitsky.
2016. {``Domain-Adversarial Training of Neural Networks.''}
\url{https://arxiv.org/abs/1505.07818}.

\leavevmode\vadjust pre{\hypertarget{ref-goodfellow2015empirical}{}}%
Goodfellow, Ian J., Mehdi Mirza, Da Xiao, Aaron Courville, and Yoshua
Bengio. 2015. {``An Empirical Investigation of Catastrophic Forgetting
in Gradient-Based Neural Networks.''}
\url{https://arxiv.org/abs/1312.6211}.

\leavevmode\vadjust pre{\hypertarget{ref-graves_connectionist_2006}{}}%
Graves, Alex, Santiago Fernández, Faustino Gomez, and Jürgen
Schmidhuber. 2006. {``Connectionist Temporal Classification: Labelling
Unsegmented Sequence Data with Recurrent Neural Networks.''} In
\emph{Proceedings of the 23rd International Conference on {Machine}
Learning - {ICML} '06}, 369--76. Pittsburgh, Pennsylvania: ACM Press.
\url{https://doi.org/10.1145/1143844.1143891}.

\leavevmode\vadjust pre{\hypertarget{ref-guo2020multisource}{}}%
Guo, Han, Ramakanth Pasunuru, and Mohit Bansal. 2020. {``Multi-Source
Domain Adaptation for Text Classification via DistanceNet-Bandits.''}
\url{https://arxiv.org/abs/2001.04362}.

\leavevmode\vadjust pre{\hypertarget{ref-He_2016}{}}%
He, Kaiming, Xiangyu Zhang, Shaoqing Ren, and Jian Sun. 2016.
{``Identity Mappings in Deep Residual Networks.''} Springer
International Publishing.
\url{https://doi.org/10.1007/978-3-319-46493-0_38}.

\leavevmode\vadjust pre{\hypertarget{ref-Tin_Kam_Ho_Nagy_2000}{}}%
Ho, Tin Kam, and G. Nagy. 2000. {``OCR with No Shape Training.''} In
\emph{Proceedings 15th International Conference on Pattern Recognition.
ICPR-2000}, 4:27--30. Barcelona, Spain: IEEE Comput. Soc.
\url{https://doi.org/10.1109/ICPR.2000.902858}.

\leavevmode\vadjust pre{\hypertarget{ref-houlsby_parameter-efficient_2019}{}}%
Houlsby, Neil, Andrei Giurgiu, Stanislaw Jastrzebski, Bruna Morrone,
Quentin de Laroussilhe, Andrea Gesmundo, Mona Attariyan, and Sylvain
Gelly. 2019. {``Parameter-Efficient Transfer Learning for NLP.''}
\url{https://doi.org/10.48550/ARXIV.1902.00751}.

\leavevmode\vadjust pre{\hypertarget{ref-hu2023llmadapters}{}}%
Hu, Zhiqiang, Yihuai Lan, Lei Wang, Wanyu Xu, Ee-Peng Lim, Roy Ka-Wei
Lee, Lidong Bing, Xing Xu, and Soujanya Poria. 2023. {``LLM-Adapters: An
Adapter Family for Parameter-Efficient Fine-Tuning of Large Language
Models.''} \url{https://arxiv.org/abs/2304.01933}.

\leavevmode\vadjust pre{\hypertarget{ref-Kirkpatrick_2017}{}}%
Kirkpatrick, James, Razvan Pascanu, Neil Rabinowitz, Joel Veness,
Guillaume Desjardins, Andrei A. Rusu, Kieran Milan, et al. 2017.
{``Overcoming Catastrophic Forgetting in Neural Networks.''}
\emph{Proceedings of the National Academy of Sciences} 114 (13):
3521--26. \url{https://doi.org/10.1073/pnas.1611835114}.

\leavevmode\vadjust pre{\hypertarget{ref-LeCun_Bengio_Hinton_2015}{}}%
LeCun, Yann, Yoshua Bengio, and Geoffrey Hinton. 2015. {``Deep
Learning.''} \emph{Nature} 521 (7553): 436--44.
\url{https://doi.org/10.1038/nature14539}.

\leavevmode\vadjust pre{\hypertarget{ref-Li_Yan_Wu_Zhu_Hu_2018}{}}%
Li, Bo, Junjie Yan, Wei Wu, Zheng Zhu, and Xiaolin Hu. 2018. {``High
Performance Visual Tracking with Siamese Region Proposal Network.''} In
\emph{2018 IEEE/CVF Conference on Computer Vision and Pattern
Recognition}, 8971--80. Salt Lake City, UT: IEEE.
\url{https://doi.org/10.1109/CVPR.2018.00935}.

\leavevmode\vadjust pre{\hypertarget{ref-liu2023improving}{}}%
Liu, Chen Cecilia, Jonas Pfeiffer, Ivan Vulić, and Iryna Gurevych. 2023.
{``Improving Generalization of Adapter-Based Cross-Lingual Transfer with
Scheduled Unfreezing.''} \url{https://arxiv.org/abs/2301.05487}.

\leavevmode\vadjust pre{\hypertarget{ref-liu2017adversarial}{}}%
Liu, Pengfei, Xipeng Qiu, and Xuanjing Huang. 2017. {``Adversarial
Multi-Task Learning for Text Classification.''}
\url{https://arxiv.org/abs/1704.05742}.

\leavevmode\vadjust pre{\hypertarget{ref-liu2022autolambda}{}}%
Liu, Shikun, Stephen James, Andrew J. Davison, and Edward Johns. 2022.
{``Auto-Lambda: Disentangling Dynamic Task Relationships.''}
\url{https://arxiv.org/abs/2202.03091}.

\leavevmode\vadjust pre{\hypertarget{ref-liu2019endtoend}{}}%
Liu, Shikun, Edward Johns, and Andrew J. Davison. 2019. {``End-to-End
Multi-Task Learning with Attention.''}
\url{https://arxiv.org/abs/1803.10704}.

\leavevmode\vadjust pre{\hypertarget{ref-mao2022unipelt}{}}%
Mao, Yuning, Lambert Mathias, Rui Hou, Amjad Almahairi, Hao Ma, Jiawei
Han, Wen-tau Yih, and Madian Khabsa. 2022. {``UniPELT: A Unified
Framework for Parameter-Efficient Language Model Tuning.''}
\url{https://arxiv.org/abs/2110.07577}.

\leavevmode\vadjust pre{\hypertarget{ref-Mathis_Breuel_2002}{}}%
Mathis, C., and T. Breuel. 2002. {``Classification Using a Hierarchical
Bayesian Approach.''} In \emph{Object Recognition Supported by User
Interaction for Service Robots}, 4:103--6. Quebec City, Que., Canada:
IEEE Comput. Soc. \url{https://doi.org/10.1109/ICPR.2002.1047410}.

\leavevmode\vadjust pre{\hypertarget{ref-Misra_Shrivastava_Gupta_Hebert_2016}{}}%
Misra, Ishan, Abhinav Shrivastava, Abhinav Gupta, and Martial Hebert.
2016. {``Cross-Stitch Networks for Multi-Task Learning.''} In
\emph{Proceedings of the IEEE Conference on Computer Vision and Pattern
Recognition}, 3994--4003.

\leavevmode\vadjust pre{\hypertarget{ref-Parisi_Kemker_Part_Kanan_Wermter_2019}{}}%
Parisi, German I., Ronald Kemker, Jose L. Part, Christopher Kanan, and
Stefan Wermter. 2019. {``Continual Lifelong Learning with Neural
Networks: A Review.''} \emph{Neural Networks} 113 (May): 54--71.
\url{https://doi.org/10.1016/j.neunet.2019.01.012}.

\leavevmode\vadjust pre{\hypertarget{ref-peng2019moment}{}}%
Peng, Xingchao, Qinxun Bai, Xide Xia, Zijun Huang, Kate Saenko, and Bo
Wang. 2019. {``Moment Matching for Multi-Source Domain Adaptation.''} In
\emph{Proceedings of the IEEE/CVF International Conference on Computer
Vision}, 1406--15.

\leavevmode\vadjust pre{\hypertarget{ref-pfeiffer2020AdapterHub}{}}%
Pfeiffer, Jonas, Andreas Rücklé, Clifton Poth, Aishwarya Kamath, Ivan
Vulić, Sebastian Ruder, Kyunghyun Cho, and Iryna Gurevych. 2020.
{``AdapterHub: A Framework for Adapting Transformers.''} In
\emph{Proceedings of the 2020 Conference on Empirical Methods in Natural
Language Processing (EMNLP 2020): Systems Demonstrations}, 46--54.
Online: Association for Computational Linguistics.
\url{https://www.aclweb.org/anthology/2020.emnlp-demos.7}.

\leavevmode\vadjust pre{\hypertarget{ref-qiu2021meta}{}}%
Qiu, Shuhao, Chuang Zhu, and Wenli Zhou. 2021. {``Meta Self-Learning for
Multi-Source Domain Adaptation: A Benchmark.''} In \emph{Proceedings of
the IEEE/CVF International Conference on Computer Vision}, 1592--1601.

\leavevmode\vadjust pre{\hypertarget{ref-Rebuffi_2017}{}}%
Rebuffi, Sylvestre-Alvise, Hakan Bilen, and A. Vedaldi. 2017.
{``Learning Multiple Visual Domains with Residual Adapters.''}
\emph{NIPS}.

\leavevmode\vadjust pre{\hypertarget{ref-rebuffi_efficient_2018}{}}%
Rebuffi, Sylvestre-Alvise, Andrea Vedaldi, and Hakan Bilen. 2018.
{``Efficient {Parametrization} of {Multi}-Domain {Deep} {Neural}
{Networks}.''} In \emph{2018 {IEEE}/{CVF} {Conference} on {Computer}
{Vision} and {Pattern} {Recognition}}, 8119--27. Salt Lake City, UT:
IEEE. \url{https://doi.org/10.1109/CVPR.2018.00847}.

\leavevmode\vadjust pre{\hypertarget{ref-Reeve_2021}{}}%
Reeve, Henry W. J., T. Cannings, and R. Samworth. 2021. {``Adaptive
Transfer Learning.''} \emph{Annals of Statistics}.
\url{https://doi.org/10.17863/cam.71165}.

\leavevmode\vadjust pre{\hypertarget{ref-rohanian2023using}{}}%
Rohanian, Omid, Hannah Jauncey, Mohammadmahdi Nouriborji, Vinod Kumar
Chauhan, Bronner P. Gonçalves, Christiana Kartsonaki, ISARIC Clinical
Characterisation Group, Laura Merson, and David Clifton. 2023. {``Using
Bottleneck Adapters to Identify Cancer in Clinical Notes Under
Low-Resource Constraints.''} \url{https://arxiv.org/abs/2210.09440}.

\leavevmode\vadjust pre{\hypertarget{ref-rothenhuxe4usler2023distributionally}{}}%
Rothenhäusler, Dominik, and Peter Bühlmann. 2023. {``Distributionally
Robust and Generalizable Inference.''}
\url{https://arxiv.org/abs/2209.09352}.

\leavevmode\vadjust pre{\hypertarget{ref-ruckle_adapterdrop_2021}{}}%
Rücklé, Andreas, Gregor Geigle, Max Glockner, Tilman Beck, Jonas
Pfeiffer, Nils Reimers, and Iryna Gurevych. 2021. {``{AdapterDrop}: {On}
the {Efficiency} of {Adapters} in {Transformers}.''} In
\emph{Proceedings of the 2021 {Conference} on {Empirical} {Methods} in
{Natural} {Language} {Processing}}, 7930--46. Online; Punta Cana,
Dominican Republic: Association for Computational Linguistics.
\url{https://doi.org/10.18653/v1/2021.emnlp-main.626}.

\leavevmode\vadjust pre{\hypertarget{ref-DBLP:journalsux2fcorrux2fRuder17a}{}}%
Ruder, Sebastian. 2017. {``An Overview of Multi-Task Learning in Deep
Neural Networks.''} \emph{CoRR} abs/1706.05098.
\url{http://arxiv.org/abs/1706.05098}.

\leavevmode\vadjust pre{\hypertarget{ref-Rusu_2016}{}}%
Rusu, Andrei A., Neil C. Rabinowitz, Guillaume Desjardins, Hubert Soyer,
James Kirkpatrick, Koray Kavukcuoglu, Razvan Pascanu, and Raia Hadsell.
2022. {``Progressive Neural Networks.''}
\url{https://arxiv.org/abs/1606.04671}.

\leavevmode\vadjust pre{\hypertarget{ref-sinha2020certifying}{}}%
Sinha, Aman, Hongseok Namkoong, Riccardo Volpi, and John Duchi. 2020.
{``Certifying Some Distributional Robustness with Principled Adversarial
Training.''} \url{https://arxiv.org/abs/1710.10571}.

\leavevmode\vadjust pre{\hypertarget{ref-vaswani2023attention}{}}%
Vaswani, Ashish, Noam Shazeer, Niki Parmar, Jakob Uszkoreit, Llion
Jones, Aidan N. Gomez, Lukasz Kaiser, and Illia Polosukhin. 2023.
{``Attention Is All You Need.''} \url{https://arxiv.org/abs/1706.03762}.

\leavevmode\vadjust pre{\hypertarget{ref-Veeramachaneni_Nagy_2003}{}}%
Veeramachaneni, Sriharsha, and George Nagy. 2003. {``Adaptive
Classifiers for Multisource OCR.''} \emph{International Journal on
Document Analysis and Recognition} 6 (3): 154--66.
\url{https://doi.org/10.1007/s10032-003-0108-x}.

\leavevmode\vadjust pre{\hypertarget{ref-wang2023comprehensive}{}}%
Wang, Liyuan, Xingxing Zhang, Hang Su, and Jun Zhu. 2023. {``A
Comprehensive Survey of Continual Learning: Theory, Method and
Application.''} \url{https://arxiv.org/abs/2302.00487}.

\leavevmode\vadjust pre{\hypertarget{ref-wang2021crosslingual}{}}%
Wang, Ziyun, Xuan Liu, Peiji Yang, Shixing Liu, and Zhisheng Wang. 2021.
{``Cross-Lingual Text Classification with Heterogeneous Graph Neural
Network.''} \url{https://arxiv.org/abs/2105.11246}.

\leavevmode\vadjust pre{\hypertarget{ref-zhao2021memory}{}}%
Zhao, Hanbin, Hui Wang, Yongjian Fu, Fei Wu, and Xi Li. 2021. {``Memory
Efficient Class-Incremental Learning for Image Classification.''}
\url{https://arxiv.org/abs/2008.01411}.

\leavevmode\vadjust pre{\hypertarget{ref-zhao2018adversarial}{}}%
Zhao, Han, Shanghang Zhang, Guanhang Wu, José MF Moura, Joao P Costeira,
and Geoffrey J Gordon. 2018. {``Adversarial Multiple Source Domain
Adaptation.''} \emph{Advances in Neural Information Processing Systems}
31.

\end{CSLReferences}

\end{document}